%% file: main.tex
\documentclass[10pt,twocolumn,letterpaper]{article}
\usepackage[pagenumbers]{wacv}      


\usepackage{soul}

\definecolor{wacvblue}{rgb}{0.21,0.49,0.74}
\usepackage[pagebackref,breaklinks,colorlinks,allcolors=wacvblue]{hyperref}

\def\wacvPaperID{2449} 
\def\confName{WACV}
\def\confYear{2026}

\title{Do Blind Spots Matter for Word-Referent Mapping? A Computational Study with Infant Egocentric Video}

\author{Zekai Shi\\
Monash University \\
{\tt\small zshi0038@student.monash.edu}
\and
Zhixi Cai\\
Monash University \\
{\tt\small zhixi.cai@monash.edu}
\and
Kalin Stefanov\\
Monash University \\
{\tt\small kalin.stefanov@monash.edu}
}

\begin{document}
\maketitle
\input{sec/0_abstract}    
\input{sec/1_intro}
\input{sec/2_relatedwork}
\input{sec/3_methods}
\input{sec/4_experiments}
\input{sec/5_results}
\input{sec/6_conclusion}
{
    \small
    \bibliographystyle{ieeenat_fullname}
    \bibliography{main}
}

\end{document}

%% file: sec/0_abstract.tex
\begin{abstract}
Typically, children start to learn their first words between 6 and 9 months, linking spoken utterances to their visual referents.
Without prior knowledge, a word encountered for the first time can be interpreted in countless ways; it might refer to any of the objects in the environment, their components, or attributes.
Using longitudinal, egocentric, and ecologically valid data from the experience of one child, in this work, we propose a self-supervised and biologically plausible strategy to learn strong visual representations.
Our masked autoencoder-based visual backbone incorporates knowledge about the blind spot in human eyes to define a novel masking strategy.
This mask and reconstruct approach attempts to mimic the way the human brain fills the gaps in the eyes' field of view.
This represents a significant shift from standard random masking strategies, which are difficult to justify from a biological perspective.
The pretrained encoder is utilized in a contrastive learning-based video-text model capable of acquiring word-referent mappings.
Extensive evaluation suggests that the proposed biologically plausible masking strategy is at least as effective as random masking for learning word-referent mappings from cross-situational and temporally extended episodes.
\end{abstract}

%% file: sec/1_intro.tex
\section{Introduction}
Language acquisition, a process by which humans learn to perceive, comprehend, and produce language, is a fundamental aspect of human development that begins as early as in utero~\cite{decasper_human_1980, Mehler1988-MEHAPO, Rebecca2000InfantArtificialLanguage}.
It is an important subfield within the study of human cognitive development.
To acquire language, humans need to learn a complex linguistic system involving a combinatorially organized sound system encompassing phonetics and phonology, an open-ended lexicon with morphological structure, and a compositional system of syntax and semantics~\cite{jackendoff1997architecture}.
No other communication system uses such a complex multilayered organization~\cite{TheFacultyofLanguageHauserNoam}.

If we view human language capacity as a finite computational system with the ability to generate an infinity of utterances, infants effectively face an intractable learning problem: based on finite evidence, they have to induce the infinity corresponding to their language.
For instance, the Providence Corpus~\cite{demuth2006Word-minimality} consists of monthly recordings of six children from age 1 to 3.
Despite the young age and small overlap of less than 20\% and about 40\% in their initial 100 and 1000 word vocabulary~\cite{hart_meaningful_1995, bornstein2004CrossLinguisticAnalysis}, children share a highly uniform grammar.

How human infants spontaneously learn this complex system remains an intriguing puzzle.
In an attempt to solve it, traditional language acquisition research involves a complex and labour-intensive process.
Researchers need to dedicate a significant amount of time and effort to observe and document the response of infants using methodologies such as eye-tracking and preferential looking studies.
The results eventually led to the creation of theoretical models designed to explain the progression of language development.
However, one limitation of these observational and experimental studies is the lack of control.
Researchers can only manipulate properties of the input data or tasks given to the children.
It is difficult to interfere with the learning mechanism and past learning experience of infants~\cite{alishahi2010computational, Monaghan2017CombiningComputationalExperimental}.
Moreover, it is also difficult and morally wrong to investigate the deprivation of a particular aspect of the learning experience that may be harmful to the growth of children.

A central challenge in modeling language acquisition is achieving both scalability and ecological validity.
One promising approach involves the use of advanced computational methods, particularly deep learning frameworks.
Through self-supervised or weakly-supervised learning paradigms, these models can take advantage of longitudinal egocentric datasets that encompass the visual and auditory input infants experience during everyday interactions.
The use of such first-person, longitudinal data from the perspective of a child is essential for preserving ecological validity, as it reflects the children's actual learning environments spanning over extended periods while constraining the model to only learn from what is available to children.
These computational approaches also allow researchers to conduct periodic testing and probing to illustrate developmental trajectories and perform thorough evaluations over hundreds and thousands of cases, which is otherwise too time-consuming to be conducted on infants.
The overall goal is to develop such a computational approach, and this paper takes a step in that direction.

We are particularly interested in the problem of grounded language acquisition and investigate the learnability of word-referent mappings using video and text.
In order to model temporal relationships, we use the state-of-the-art VideoMAEv2~\cite{videomae} capable of learning strong visual representations from video.
However, the standard random masking strategy in VideoMAEv2 is not ecologically valid and difficult to justify.
Therefore, we propose a biologically plausible masking approach based on the location and size of the eyes' blind spots.
A multimodal contrastive learning-based video-text model utilizes the video representations learned with this novel masking strategy to pull together temporally aligned video-text pairs and push apart unaligned pairs.
Our contribution is four-fold:

\begin{itemize}
    \item We propose a biologically plausible masking for self-supervised learning of visual representations from video.
    \item We evaluate the proposed masking on the downstream task of word-referent mapping using an egocentric longitudinal dataset from a child's perspective.
    \item We further evaluate the generalization to novel, unseen developmentally relevant datasets.
    \item We curate a new labeled video egocentric longitudinal dataset from a child's perspective.
\end{itemize}

%% file: sec/2_relatedwork.tex
\begin{table}[t]
    \setlength{\tabcolsep}{1pt}
    \footnotesize
    \centering
    \begin{tabular}{llllccc}
        \toprule
        \textbf{Datasets} & \textbf{Type} & \textbf{\#P} & \textbf{Length} & \textbf{Audio} & \textbf{Transcript} & \textbf{Gyroscope} \\
        \midrule
        BV-Home~\cite{long2024babyviewdatasethighresolutionegocentric} & Infant & 28 & 433 & \checkmark & \checkmark & \checkmark \\
        BV-Preschool~\cite{long2024babyviewdatasethighresolutionegocentric}& Child & 39 & 63 & \checkmark & \checkmark & \checkmark \\
        Ego-SingleChild~\cite{long2024babyviewdatasethighresolutionegocentric} & Infant & 1 & 47 &  \checkmark &  \checkmark & \\
        SAYCam~\cite{sullivan2021SAY} & Infant & 3 & 476 & \checkmark & \checkmark & \\
        Homeview~\cite{FAUSEY2016faces2hands} & Infant & 101 & 500 & \checkmark &  & \\
        \bottomrule
    \end{tabular}
    \caption{\textbf{Egocentric datasets from a child's perspective.} \#P denotes the number of participants, and the length is in hours.}
    \label{tab:egocentricDataset}
\end{table}

\section{Related Work}
\noindent\textbf{Computational Modeling.}
There are many computational approaches to language acquisition in the literature.
Pursuit is a statistical computational model that follows a hypothesis testing approach~\cite{stevens2017Pursuitofwordmeanings}.
Bayesian modeling is also used for language acquisition~\cite{ABEND2017BootstrappingBay}.
For deep learning approaches, many use self-supervised learning and examine the learnability problem from either unimodal text~\cite{qin2024systematicinvestigationlearnabilitysingle}, audio~\cite{lavechin2025LanguagePhonetic}, video~\cite{orhan2024selfsupervisedlearningvideorepresentations}, and image~\cite{orhan2020selfsupervisedlearningeyeschild, DAVIDSON2024spatialrelationcategorization, orhan2024withoutstronginductive} inputs, or multimodal learning with image and text~\cite{vong2024groundedlanguageacquisition}, or with image and audio~\cite{KHORRAMI2025LAIA}.
Comparing models pretrained on SAYCam~\cite{sullivan2021SAY} and Imagenet~\cite{imagenet}, \citet{DAVIDSON2024spatialrelationcategorization} examined the learnability of spatial relations such as above, below, containment, and between.
\citet{sheybani2023CurriculumInfant} used a generative model (VideoMAE~\cite{videomae}), a predictive model (modified I-JEPA~\cite{I-JEPA} for video), and a contrastive model (modified SimCLR~\cite{SIMCLR} for video) for curriculum learning on the Homeview dataset~\cite{FAUSEY2016faces2hands}.
The proposed BlindSpotMAE examines the learnability of word-referent mappings and employs contrastive learning with video-text pairs using egocentric data from the perspective of a child.
The model is also examined on spatio-temporal understanding tasks using a developmentally relevant benchmark.

\begin{table}[t]
    \setlength{\tabcolsep}{1pt}
    \footnotesize
    \centering
    \begin{tabular}{llccccc}
        \toprule
        \textbf{Benchmarks} & \textbf{Modality} & \textbf{Ego} & \textbf{Developmental} & \textbf{Lex} & \textbf{Syn} & \textbf{Sem} \\
        \midrule
        Toybox~\cite{wang2018Toybox} & video-text & \checkmark & \checkmark & \checkmark & & \\
        Labeled-S~\cite{orhan2020selfsupervisedlearningeyeschild} & image-text & \checkmark & \checkmark & \checkmark & & \\
        Zorro~\cite{huebner-etal-2021-babyberta} & text & & \checkmark & & \checkmark & \\
        EgoObjects~\cite{zhu2023egoobjects} & video-text & \checkmark & & \checkmark & & \\
        DEVBENCH~\cite{tan2024devbenchmultimodaldevelopmentalbenchmark} & image-text & & \checkmark & \checkmark & \checkmark & \checkmark \\ 
        ModelVsBaby~\cite{sheybani_modelvsbaby_2024} & image-text & & \checkmark & \checkmark & & \\
        \midrule
        Video Labeled-S & video-text & \checkmark & \checkmark & \checkmark & & \\
        \bottomrule
    \end{tabular}
    \caption{\textbf{Benchmarks for evaluation.} Ego: Egocentric; Lex: Lexicon; Syn: Syntax; Sem: Semantics.}
    \label{tab:benchmarkEvaluation}
\end{table}

\noindent\textbf{Datasets and Benchmarks.}
SAYCam~\cite{sullivan2021SAY} is a video corpus of 415 hours of egocentric recordings from three infants aged 6-32 months.
Babyview dataset~\cite{long2024babyviewdatasethighresolutionegocentric} is another video corpus containing 493 hours of high-resolution egocentric videos.
Data are collected from 28 families and a preschool classroom.
Homeview dataset~\cite{FAUSEY2016faces2hands} is a dataset used in \cite{sheybani2023CurriculumInfant}.
It contains 500 hours of head camera recordings from 101 infants.
Toybox~\cite{wang2018Toybox} is a developmentally relevant video dataset with 12 object categories.
Each category contains 30 unique toy objects that undergo 12 spatial transformations.
Devbench~\cite{tan2024devbenchmultimodaldevelopmentalbenchmark} is a multimodal developmental benchmark with the focus on language learning and similarity between model and human responses.
It contains seven tasks covering semantic, syntactic, and lexical tasks suitable for a variety of age groups.
ModalVSBaby~\cite{sheybani_modelvsbaby_2024} is an out-of-distribution object benchmark.
It also provides responses from 2-year-old children.
Zorro~\cite{huebner-etal-2021-babyberta} is a grammar test suite, and EgoObjects~\cite{zhu2023egoobjects} contains 368 categories with bounding boxes.
Our curated video Labeled-S dataset aims to address the gap of a lack of developmentally relevant egocentric benchmarks for evaluating video and video-text models.
See \cref{tab:egocentricDataset} for an overview of available egocentric developmental datasets and \cref{tab:benchmarkEvaluation} for benchmarks and datasets for evaluation.

\begin{figure*}[t]
    \centering
    \includegraphics[width=0.9\linewidth]{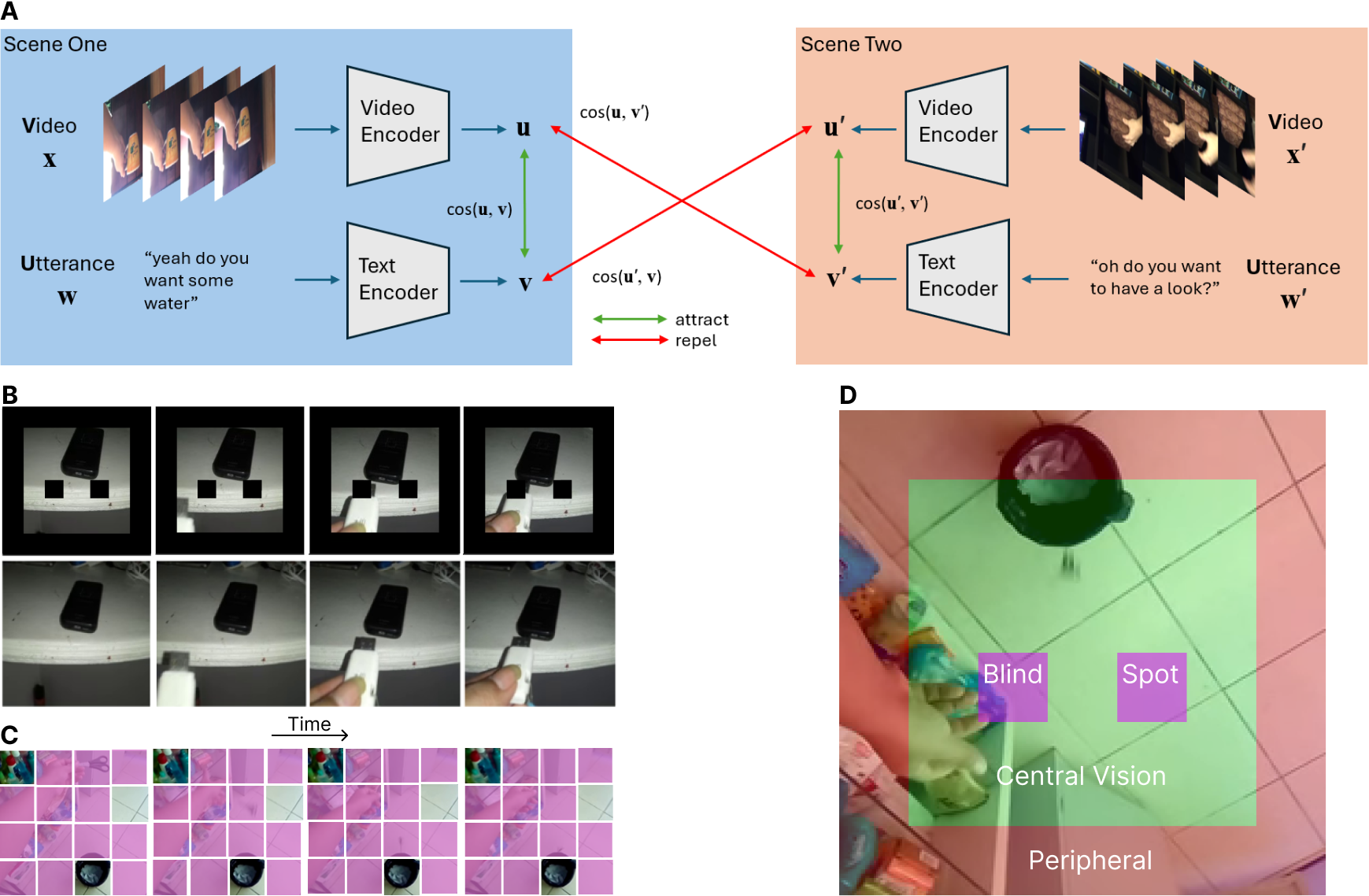}
    \caption{\textbf{Model architecture, masking strategy, and spatio-temporal attention.} (\textbf{A}) Video and utterance are embedded using their corresponding encoder. Video-utterance pairs in the same scene are brought together, and video-utterance pairs in different scenes are pulled away. (\textbf{B}) An example of blind spot masking, where black squares are masked patches, and the second row shows the original frames. (\textbf{C}) An example of tube masking where images with purple blocks are masked patches, and the position of unmasked patches is chosen at random for each clip, but constant for all frames of a clip. (\textbf{D}) A visualization of different regions of blind spot masking, assuming videos are captured with a 109$\times$70 field-of-view camera. Example frames were taken from SomethingSomethingV2\cite{goyal2017somethingsomethingvideodatabase}}
    \label{fig:method-overview}
\end{figure*}

\noindent\textbf{Video Representation Learning.}
Models like S3D~\cite{Xie2018S3D} use convolutional layers for video understanding.
Most video foundation models are trained with a vision transformer (ViT~\cite{dosovitskiy2020vit}) backbone.
VideoMAE~\cite{videomae} uses tube masking with a high masking ratio of 90\%.
VideoMAEv2~\cite{wang2023videomaev2} extends upon VideoMAE by training a billion-level vision transformer with masking at the convolutional neural network decoder.
InternVideo2~\cite{wang2024internvideo2} is a multimodal video foundation model with 6 billion parameters, ViT, and a three-stage training scheme.
VideoMamba~\cite{li2024videomamba} is a video model that is purely based on a selective state space model inspired by Mamba~\cite{mamba}.
It achieved strong performance with less computing for long video understanding.
VideoMamba is trained using unmasked token alignment with a teacher inspired by UMT~\cite{Li2023UMT}.
While previous work uses variations of random masking, we shift from those standard approaches by enforcing ecologically valid masking.

%% file: sec/3_methods.tex
\section{Method}
An overview of the multimodal model used for word-referent mapping is shown in~\cref{fig:method-overview}.

\noindent\textbf{Input.} The model takes pairs of temporally aligned video clips $X \in \mathbb{R}^{3 \times T \times H \times W}$ consisting of $T$ RGB video frames of size $H \times W$ and child-directed utterances $W \in \mathbb{Z}^L$ consisting of a sequence of integers representing an utterance containing $L$ words.

\noindent\textbf{Video Encoder.}
The video encoder $f_\theta$ is a VideoMAEv2~\cite{wang2023videomaev2} with ViT~\cite{dosovitskiy2020vit} backbone. 
For pretraining, VideoMAEv2 uses a high masking ratio of 90\% with random tube masking.
It employs cube patch embedding that uses a \texttt{conv3d} to convert a video patch of size $2\times16\times16$ into one token embedding.
Each token is then added with a positional embedding.
The unmasked token is then passed to the ViT encoder with vanilla joint space-time attention~\cite{gberta2021spaceTimeAttention}.
A shallow decoder consisting of CNN layers was used to reconstruct the image from the encoder and learnable mask token. 
During pretraining, the mean squared error (MSE) loss is formulated between the normalized masked tokens and their reconstructed counterparts in the pixel space.

\begin{equation}
    L_{\text{MSE}} = \frac{1}{|\mathcal{M}|} \sum_{i \in \mathcal{M}} \left(I_i - \hat{I}_i \right)^2
\end{equation}
where $I_i$ and $\hat{I}_i$ are the input and reconstructed pixel values of token $i$, respectively, and $\mathcal{M}$ is the set of masked tokens. 
During multimodal training, the decoder of VideoMAEv2 is removed and replaced with a linear layer that projects the latent representation to a $d$-dimensional vector $\boldsymbol{u}$.


\noindent\textbf{Blind Spot Masking.}
Random tube masking masks random patches of an input clip, which is different from how humans process vision.
In this work, we aim to train vision models in a more biologically plausible way.
Inspired by humans' perceptual filling of the blind spots~\cite{1992BlindSpot}, we propose blind spot masking, a masking that mimics how humans will perceive the world in an egocentric view.
Human blind spots subtend $5^\circ \times 7^\circ$ visual angle and are centered about $15^\circ$ temporally.
In our formulation, blind spot masking assumes a form of binocular vision as we mask the blind spots from both eyes and the periphery for computational efficiency.

To calculate the blind spot masking, we take into consideration the field-of-view (FOV) of the camera that captures the video.
We assume that the child will see the clip in a similar FOV.
For example, in the case of SayCam~\cite{sullivan2021SAY}, the camera has an FOV of $109^\circ \times 70^\circ$.
Central vision covers a $60^\circ \times 60^\circ$ FOV~\cite{visualFieldCentral}, and we calculate the central vision region using the video resolution.
Let $h$, $w$ be the height and width of the video, $FOV_{vh}$, $FOV_{vw}$ be the FOV of the egocentric camera that captures the video, $FOV_{ph}$, $FOV_{pw}$ be the FOV covered by humans' central vision.
We calculate $h_c$, $w_c$, the height and width of central vision in pixels:

\begin{equation}
    h_c = h / FOV_{vh} * FOV_{ph}
\end{equation}
\begin{equation}
    w_c = w / FOV_{vw} * FOV_{pw}
\end{equation}

We calculate the size of the blind spots in a similar way.
We slightly alter the size of the calculated central vision so that it could fit in the $16\times16$ patch of the model.
The actual masking is then computed by masking everything that is outside of the central vision and everything that is covered by the blind spots. See section D of \cref{fig:method-overview} for an example of how we define each region assuming a $ FOV_{vh}$ of $109^\circ$ and $FOV_{vw}$ of $70^\circ$.



\noindent\textbf{Text Encoder.}
Like~\cite{vong2024groundedlanguageacquisition}, the text encoder $f_\phi$ consists of a single trainable embedding layer that maps each word token into a $ d$-dimensional vector.
For utterances of many words, a single embedding is obtained by computing the mean of the embedding layer output for each word in the utterance.

\noindent\textbf{Contrastive Loss.}
Given a batch of size $B$, inter-modal contrastive InfoNCE~\cite{oord_representation_2019} loss is used to pull together co-occurring text and video representations and push away mismatched ones.
Let $\boldsymbol{u}$ and $\boldsymbol{v}$ represent the embedded videos and texts inside a batch, $u_i$ and $v_i$ be the $i$th embedded video text pair in the batch.
The loss will be the symmetric loss:

\begin{equation}
    L = \text{InfoNCE}(\boldsymbol{u},\boldsymbol{v}) + \text{InfoNCE}(\boldsymbol{v},\boldsymbol{u})
\end{equation}
where InfoNCE$(\boldsymbol{u},\boldsymbol{v})$ is the video-text contrastive loss and the InfoNCE$(\boldsymbol{v},\boldsymbol{u})$ is the text-video contrastive loss.
Specifically, the video-text contrastive loss is given by:

\begin{equation}
    \text{InfoNCE}(\boldsymbol{u},\boldsymbol{v}) = -{1\over{B}}\sum_{i}^B\log{\exp(u_i^Tv_i/\tau)\over\sum_{j=1}^B\exp(u_i^Tv_j/\tau)}
\end{equation}

\begin{table}[t]
    \setlength{\tabcolsep}{5pt}
    \centering
    \begin{tabular}{llll}
        \toprule
        \textbf{Length} & \textbf{Sim} & \textbf{Stride} & \textbf{\#Videos} \\
        \midrule
        4  & 0 & 0 & 52657 \\
        4  & 0.99 & 0 & 24421 \\
        4  & 0.99 & 4 & 7285 \\
        2.3* & 0 & 0 & 52657 \\
        2.3* & 0.99 & 0 & 24421 \\
        2.3* & 0.99 & 4 & 7285 \\
        \bottomrule
    \end{tabular}
    \caption{\textbf{Video Labeled-S dataset statistics.} Length and stride are in seconds. The 2.3-second length is an estimate of a fixed number of 68 frames sampled from each clip. 68 frames are chosen to allow video models that use 16-frame inputs to sample every fourth frame for diversity. Videos in SAYCam are captured at 30fps, but the actual fps fluctuates between 20 and 30 due to constraints of the camera. Sim is the minimum similarity threshold.}
    \label{tab:videoLabeledSStats}
\end{table}

\begin{table*}[t]
\centering
\begin{tabular}{ccccccccc}
\toprule
\textbf{Model} & \textbf{Data } & \multicolumn{2}{c}{\textbf{Labeled-S}} & \multicolumn{2}{c}{\textbf{Video Labeled-S}} & \multicolumn{2}{c}{\textbf{Toybox Object}} \\
\cmidrule(lr){3-4} \cmidrule(l){5-6} \cmidrule(l){7-8}
& & \textbf{Acc@1} & \textbf{Acc@5} & \textbf{Acc@1} & \textbf{Acc@5} & \textbf{Acc@1} & \textbf{Acc@5} \\
\midrule
DINO ResNext-50~\cite{caron2021emerging}  & Child-S & 73.332 & 96.658 & 74.677 & 96.029 & 73.238 & 96.623 \\
DINO~\cite{caron2021emerging}  & Child-S & 72.760 & 95.566 & 74.231 & 96.427 & 81.054 & 98.480 \\
\midrule
VideoMAEv2~\cite{wang2023videomaev2}  & Child-S & \textbf{58.815} & \textbf{90.781} & 56.300 & 88.911 & 33.437 & 78.212 \\
BlindSpotMAE-200x135  & Child-S & 56.576 & 88.833 & 56.852 & \ul{89.116} & \textbf{41.244} & \textbf{83.866} \\
BlindSpotMAE-Center  & Child-S & 55.410 & 88.331 & \textbf{60.842} & \textbf{91.008} & \ul{41.088} & \ul{84.119} \\
BlindSpotMAE-109x70  & Child-S & \ul{57.306} & \ul{89.715} & \ul{57.150} & 89.057 & 36.528 & 80.674 \\
BlindSpotMAE-NoPeripheral & Child-S & 45.357 &82.130 &47.290 &83.594 &27.936 &71.598 & \\
\bottomrule
\end{tabular}
\caption{\textbf{Linear Decoding Accuracy} for object understanding. Two DINO Models are image models. All models use ViT-B as their backbone, except the first DINO model, which uses ResNext50. For video models, the highest number in each column is highlighted in \textbf{bold}, and the second-highest is \underline{underlined}.}
\label{tab:linear_decode_obj}
\end{table*}

\begin{table}[t]
    \setlength{\tabcolsep}{5pt}
    \centering
    \begin{tabular}{cccccccc}
        \toprule
        \textbf{Length} & \textbf{Sim} & \textbf{Stride} & \textbf{Acc@1} & \textbf{Acc@5} \\
        \midrule
        4  & 0 & 0  & 50.294 & 86.110\\
        4  & 0.99 & 0 & 56.126 & 88.666 \\
        4  & 0.99 & 120 & 47.015 & 83.824 \\
        2.3 & 0 & 0 & 50.534 & 86.210 \\
        2.3 & 0.99 & 0 & 57.150 & 89.057 \\
        2.3 & 0.99 & 120 & 48.455 & 85.688 \\
        \bottomrule
    \end{tabular}
    \caption{\textbf{Linear Decoding Accuracy for Video Labeled-S.} Length and stride are in seconds. The 2.3-second length is an estimate of a fixed number of 68 frames sampled from each clip. 68 frames are chosen to allow video models that use 16-frame inputs to sample every fourth frame for diversity. Videos in SAYCam are captured at 30fps, but the actual fps fluctuates between 20 and 30 due to constraints of the camera. Sim is the minimum similarity threshold. All results were obtained with BlindSpotMAE-109$\times$70.}
    \label{tab:videoLabeledSAcc}
\end{table}

\section{Video Labeled-S Dataset}
Currently, there is a lack of developmentally relevant video evaluation datasets.
To address this gap, we aim to contribute to existing evaluation resources by curating a labeled video dataset.
Our approach begins by leveraging the state-of-the-art image model DINOv2~\cite{oquab2023dinov2} with a ViT-B/14 backbone to compute latent representations for all images in the Labeled-S dataset \cite{orhan2020selfsupervisedlearningeyeschild}, as well as for every 30th frame extracted from videos in the Child-S dataset \cite{sullivan2021SAY}.
We then calculate the cosine similarity between each image in Labeled-S and all sampled frames from the Child-S videos.
For each image in Labeled-S, we identify the frame with the highest cosine similarity and designate it as the corresponding match.
A video clip is subsequently sampled around this matched frame.

In practice, we experimented with multiple clip lengths and applied a filtering step, retaining only clips that met a minimum cosine similarity threshold.
We also experimented with having a stride requirement between the centers of extracted clips to prevent overlaps in frames between adjacent sampled clips. 
See \cref{tab:videoLabeledSStats} for an overview of statistics on the dataset under various settings.

%% file: sec/4_experiments.tex
\section{Experiments}
We performed evaluations, probing the video encoder performance on classification tasks with Toybox \cite{wang2018Toybox} and Labeled-S \cite{orhan2020selfsupervisedlearningeyeschild} datasets.
We also evaluate the multimodal model with text-video retrieval.
Image inputs to the video model are arranged as clips by stacking the image 16 times.

\subsection{Datasets}
\noindent\textbf{SAYCam}~\cite{sullivan2021SAY} is a longitudinal dataset of head-mounted camera recordings from the perspective of three children (S, A, and Y).
For each child, videos were recorded for approximately two hours per week.
It was also strongly encouraged that one hour of the recording occur at a fixed time, with the other hour to occur at a random time.
In this work, the video encoder is first pretrained on recordings from child S only.
Specifically, there are 194 hours of video recordings from child S.
Child S is 6 months old at the first recording and 30 months old at the last recording.
Recordings are at 480p with 30fps.

\noindent\textbf{Labeled-S} is a labeled image dataset derived from SAYCam, where all images come from child S.
The details for the process behind curating the dataset are outlined in \cite{orhan2020selfsupervisedlearningeyeschild}.
The dataset contains 58K frames from 26 classes.
The class mainly comes from the field ``object being looked at'' in the manual transcript of recordings from child S.
All images in the dataset also have the resolution $224\times224$.

\noindent\textbf{Toybox}~\cite{wang2018Toybox} is a video dataset of egocentric views of 12 classes of objects that undergo 12 transformations.
Videos are recorded with 1080p at 30fps.
Each class has 30 objects, and the dataset contains roughly 2.3M frames, which equate to 21 hours of video.
Objects in the dataset mainly consist of child toys representing the particular class (e.g., toy giraffe, toy helicopter).
Transformations include rotation in the positive and negative $x$, $y$, and $z$ axes, translation in the $x$, $y$, and $z$ axes, and a hodgepodge motion.
Two special transformations are `absent', where the object is not in the video, and `present', where the object remains still in the video. 
We can extract both the object and transformation labels. 
We denote Toybox Object as the dataset used for object classification and Toybox Transformation as the dataset used for classifying transformations.

\subsection{Implementation Details} 
We employ the AdamW~\cite{loshchilov_decoupled_2019} optimizer for all training phases. 
For video encoder pretraining, we set the learning rate to $1.5\times10^{-4}$ with a batch size of 16, and a weight decay of 0.05.
For multimodal contrastive learning, we use a learning rate of $1\times10^{-4}$, a batch size of 16, a weight decay of 0.1, and a fixed temperature of 0.07.

\noindent\textbf{Pretraining.} Video encoders are pretrained following standard VideoMAEv2 pretraining procedure~\cite{wang2023videomaev2}.
They are trained on videos from child S only.
These recordings were split into 8-second chunks as they can often be over 20 minutes long.
During pretraining, every four consecutive frames were sampled, and each clip contains 16 frames, covering approximately 2.1 seconds of video time.
Data augmentation for videos involves a random multiscale crop followed by normalization with the ImageNet~\cite{imagenet} mean and standard deviation.

\noindent\textbf{Pretraining Model Variations.}
We trained four BlindSpotMAE models with different settings. Human binocular vision covers around a 200x135 FOV~\cite{Grove2013BinocularVision, doi:10.2352/J.Percept.Imaging.2018.1.1.010505}. 
BlindSpotMAE-200x135 assumes that the video clip covers a similar FOV as the human binocular vision during the calculation of blind spot masking. 
BlindSpotMAE-109x70 uses the same FOV as the camera used in the Child S data for calculating the blind spot masking. 
BlindSpotMAE-Center uses the center crop of the input clip by incorporating the assumption that the camera is generally pointing in the direction that the child is fixating on. This model also assumes a 200 $\times$ 135 FOV.
BlindSpotMAE-NoPeripheral examines the necessity of masking the peripheral vision by only masking the blind spots and unmasking all peripheral regions.

\begin{table*}[t]
    \centering
    \begin{tabular}{cccc}
        \toprule
        \textbf{Model} & \textbf{Data}  & \multicolumn{2}{c}{\textbf{Toybox Transformation}} \\
        \cmidrule(lr){3-4}
        & & \textbf{Acc@1} & \textbf{Acc@5} \\
        \midrule
        DINO ResNext-50~\cite{caron2021emerging} & Child-S  & 28.998 & 78.384 \\
        DINO~\cite{caron2021emerging} &  Child-S  & 30.387 & 79.907 \\
         \midrule
        VideoMAEv2~\cite{wang2023videomaev2} & Child-S  & \textbf{60.438} & \textbf{93.140} \\
        BlindSpotMAE-200x135 & Child-S & 49.049 & 88.695 \\
        BlindSpotMAE-Center  & Child-S  & 43.426 & 85.446 \\
        BlindSpotMAE-109x70  & Child-S & \ul{58.527} & \ul{91.675} \\
        BlindSpotMAE-NoPeripheral & Child-S & 55.598 & 90.396 \\
        
        \bottomrule
    \end{tabular}
    \caption{\textbf{Linear Decoding Accuracy} for spatial understanding. Two DINO Models are image models. All models use ViT-B as their
backbone, except the first DINO model, which uses ResNext50.For video models, the highest number in each column is highlighted in \textbf{bold}, and the second-highest is \underline{underlined}. }
    \label{tab:linear_decode_act}
\end{table*}

\noindent\textbf{Multimodal Training.} The video encoder is frozen during the multimodal learning phase.
We follow the same data curation procedure as~\cite{vong2024groundedlanguageacquisition} to get a training set of images and child-directed utterance pair.
During training, we concatenate the same image 16 times to form a clip for multimodal model with video encoders. 

\subsection{Evaluation Details}
We compare the performance of our pretrained video models against image models pretrained in~\cite{orhan2024withoutstronginductive} on the following tasks. All video and image models are pretrained with the child S data from SAYCam \cite{sullivan2021SAY}. We also compare their performance when employed on a contrastive learning architecture. All models except the two image DINO models are trained by us.
All DINO models are image-based models taken \cite{orhan2024withoutstronginductive}.
Since Labeled-S is an image dataset and Toybox is a video dataset, in order to evaluate an image model on videos, we evaluate the image model on each image of a clip to get the mean of the logits and obtain a prediction.

\noindent\textbf{Object Classification.}
To examine the ability of the video encoder on object classification, we evaluate its performance on Labeled-S, Video Labeled-S, and Toybox Object datasets. 
Labeled-S is divided into a 45\% training, 5\% validation, 50\% testing split. 
Evaluation on all image datasets consists of a simple normalization with ImageNet mean and standard deviation. 
For evaluation on video datasets, we use the VideoMAEv2 video loading strategy where validation uses the center crop, training uses spatial sampling, random erasing, color jittering, and rand augment policy~\cite{Cubuk2020RandAugment}. 
Both Video Labeled-S and Toybox Object employ an 80\% training, 10\% validation, and 10\% testing split. 
For Video Labeled-S, we report the performance of different models on the 2.3 seconds, 0 stride, 0.99 minimum cosine similarity version of the dataset.

\noindent\textbf{Spatio-Temporal Understanding.}
We evaluate the spatio-temporal understanding of the pretrained video autoencoder on the Toybox Transformation dataset using the same VideoMAEv2 video loading strategy with an 80\% training, 10\% validation, and 10\% testing split.

\noindent\textbf{CVCL Evaluation.}
We also evaluate the performance of the video encoder using the CVCL~\cite{vong2024groundedlanguageacquisition} evaluation trials.
Each trial consists of four images, and three of them are of a foil category.
The model needs to predict the highest probability for the right image given the category label.
This evaluation strategy mimics the modern PPVT (Peabody Picture Vocabulary Test)~\cite{dunn_2007_peabody} test that measures children's vocabulary skills. 

\noindent\textbf{Video-Text Retrieval.} To examine the learned word–referent mappings, we evaluated video–text retrieval on the Labeled-S dataset. 
The evaluation protocol was identical to that used in the CVCL evaluation, except that it was applied to a multimodal model.
For each trial, the model predicted the referent by selecting the image with the highest cosine similarity to the given word token.

%% file: sec/5_results.tex
\section{Results}


\subsection{Object Classification}
\label{subsec:objectClassificationResult}
\cref{tab:linear_decode_obj} reports the linear decoding accuracy for object classification. 
There exists a systematic gap between the image and video models as a whole on the object classification task. 
The best image model (DINO with ResNext-50 backbone) performs +14.52\% compared to VideoMAEv2 with ViT-B backbone trained using child S data in terms of Top-1 accuracy. 
Within the video models, we observe that models with blind spot masking achieve performance comparable to VideoMAEv2. Notably, some models, such as BlindSpotMAE-200$\times$135, surpass VideoMAEv2, achieving a 7.807\% improvement on the Toybox Object dataset. Interestingly, BlindSpotMAE-109$\times$70 employs a substantially lower masking ratio (53\%) compared to the standard VideoMAE masking ratio of 90\%. Despite previous reports of reduced accuracy with low masking ratios \cite{videomae}, BlindSpotMAE-109$\times$70 performs at a level comparable to VideoMAEv2 and consistently outperforms it on object classification tasks with video input. This highlights the generalisability of representations learned using a biologically inspired blind spot masking strategy. 

In contrast, BlindSpotMAE with no peripheral masking performs worse across all object classification datasets regardless of the modality. We hypothesize that this is due to the extremely small masking ratio of 6\% and potential information leakage. Furthermore, the reconstruction task is quite simple, as the model only needs to reconstruct the blind spot. These findings demonstrate the importance of including peripheral masking for learning robust video representations with a blind spot masking strategy.

Finally, \cref{tab:videoLabeledSAcc} shows the linear decoding accuracy achieved by  BlindSpotMAE-109$\times$70 across different settings of the Video Labeled-S dataset. Through this experiment, we aim to provide baseline performances on various settings of the dataset. The performance drops as stricter filtering are applied, we hypothesise that this is caused by the decrease in the number of training samples. The dataset might be helpful for evaluating the data efficiency of models. 

\begin{table}[t]
    \centering
    \begin{tabular}{ccccc}
        \toprule
        \textbf{Model}  & \textbf{Data} & \textbf{Accuracy} \\
        \midrule
         DINO ResNext-50~\cite{caron2021emerging}   & Child-S&  81.10 \\
        DINO~\cite{caron2021emerging} & Child-S & 90.50 \\
        \midrule
        VideoMAEv2~\cite{wang2023videomaev2} &  Child-S & \textbf{82.81} \\
        BlindSpotMAE-200x135 &  Child-S  & 81.50 \\
        BlindSpotMAE-Center &  Child-S & 81.50 \\
        BlindSpotMAE-109x70 &  Child-S  & \ul{82.41} \\
        BlindSpotMAE-NoPeripheral & Child-S &73.95 \\
        \bottomrule
    \end{tabular}
    \caption{\textbf{Labeled-S Evaluation Trials Accuracy}. Evaluate linear decoding models on CVCL evaluation trials where the model needs to find the image of the correct class given one correct image and three other images from foil categories.}
    \label{tab:linear_decode_eval}
\end{table}

\subsection{Spatio-Temporal Understanding}
\cref{tab:linear_decode_act} reports the linear decoding accuracy for spatio-temporal understanding. The best performing video encoder significantly outperforms the image encoder on Toybox Transformation dataset (+30.053\% in terms of Top-1 accuracy), suggesting its representations are better suited to capture spatial transformations and integrate information over multiple time steps.
In addition to that, since the Toybox dataset contains transformations of 420 different instances of objects, the ability of the video encoder to perform well on the dataset highlights its capability to generalize the spatio-temporal pattern to a diverse range of objects that are unseen during the pretraining phase.
BlindSpotMAE-109x70 again is able to achieve similar performance as the VideoMAEv2 model, suggesting that spatio-temporal understanding is indeed learnable even if we constrain the model with the same mask positions.
BlindSpotMAE based on 200$\times$135 FOV with center crop performs the worst out of the four video models. 
We hypothesise that it might be caused by the high masking ratio and the overly difficult reconstruction task when the model can only see the center 10\% of input clip. 

\subsection{CVCL Evaluation}
Following \cite{vong2024groundedlanguageacquisition}, we also evaluate linear decoding models on the choice of one out of 4 video-text evaluation trials, as shown in \cref{tab:linear_decode_eval}.
From this test, it appears the gap between image and video models is reduced, highlighting the video models' capability to select the right class from a limited set of foil classes.
While video models may perform worse at direct object classification, they nonetheless demonstrate a comparable ability to infer the correct class when constrained to a selection of candidate classes. The performance pattern between video autoencoders in the evaluation trial is consistent with that observed in the linear decoding for the Toybox Transformation dataset, BlindSpotMAE-109x70 performing at a level very similar to that of the VideoMAEv2 model.


\begin{figure*}[t]
    \centering
    \includegraphics[width=0.9\linewidth]{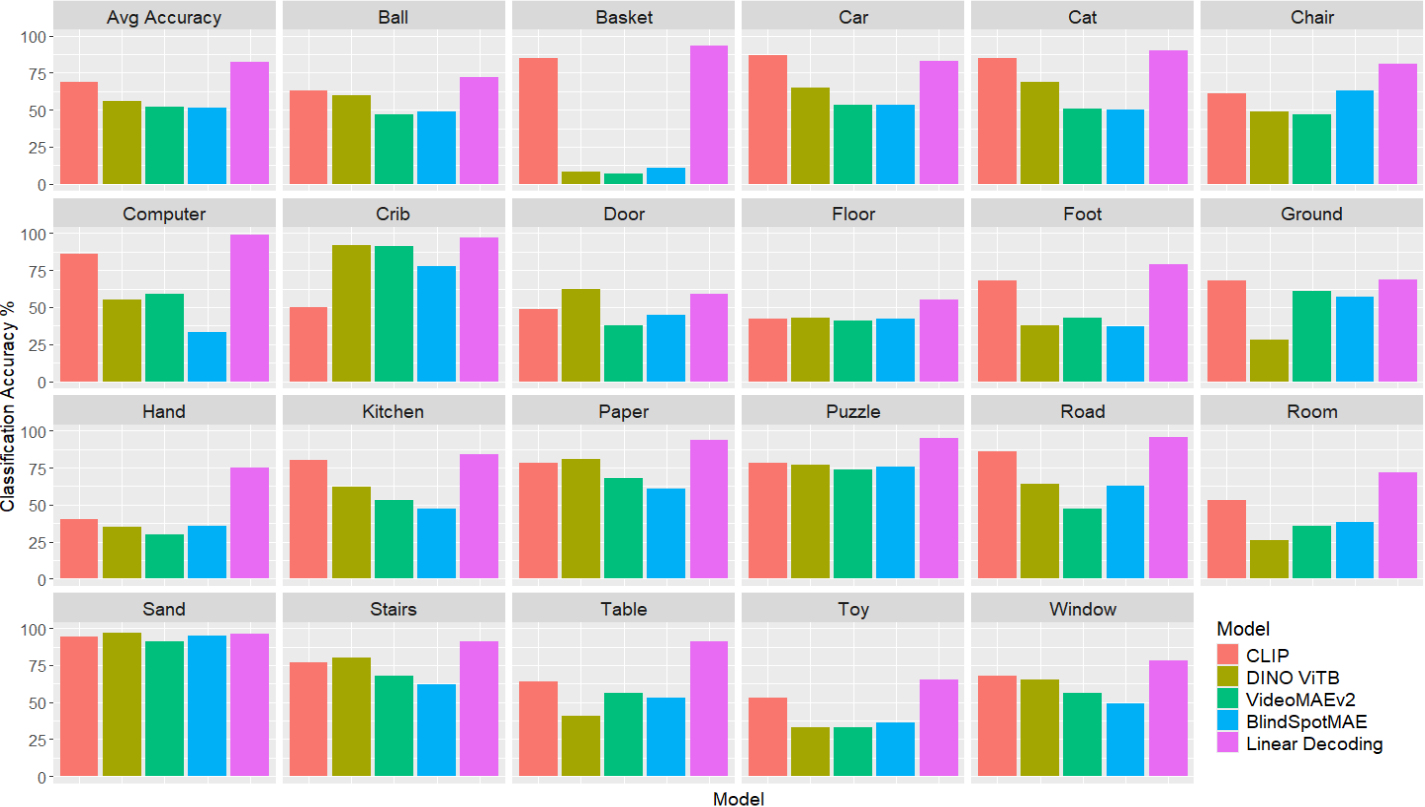}
    \caption{\textbf{Classification Accuracy for Labeled-S dataset}. CLIP, DINO ViTB, BlindSpotMAE, and VideoMAEv2 all refer to multimodal models. VideoMAEv2 and BlindSpotMAE-109$\times$70 are used as a pretrained video encoder with ViT-B backbone. Linear decoding refers to evaluating only the BlindSpotMAE video encoder with a linear head on the CVCL evaluation task as shown in \cref{tab:linear_decode_eval}}
    \label{fig:multilabel_result}
\end{figure*}

\subsection{Multimodal Text-Video Retrieval}
\cref{fig:multilabel_result} shows the accuracies of different models on the CVCL retrieval task.
CLIP~\cite{radford2021learningtransferablevisualmodels} is an image-text contrastive learning model that was trained on 400 million image-text pairs. Linear Decoding is the accuracy achieved by BlindSpotMAE-109x70 in \cref{tab:linear_decode_eval} in the CVCL evaluation trial. DINO ViTB, VideoMAEv2, BlindSpotMAE refers to the pretrained frozen visual encoder used during contrastive learning. DINO ViTB and CLIP are all image-text models. 

Both VideoMAEv2 and BlindSpotMAE visual encoders perform similarly, with the VideoMAEv2 model achieving an average accuracy that is 0.73\% higher than the BlindSpotMAE model.
Their accuracies are also closely aligned within individual classes, suggesting the learning of more uniform representations. 
In contrast, the linear probe model achieves substantially better performance (+13.55\%) compared to the CLIP model, despite CLIP being trained on a much larger and more diverse dataset. This improvement reflects the advantage of supervision on in-distribution samples.
%

The image-based multimodal model with DINO ViT-B visual encoder also demonstrates a similar performance pattern across classes compared to the video encoders. 
Despite evaluating the video encoder on stacked still images, the multimodal model still achieves competitive performance (-3.639\%) on the retrieval task compared to the image-text model. 
This result is in stark contrast to the large gap between video and image models in \cref{subsec:objectClassificationResult}. 
It appears that having language supervision allows the model to utilize the representation from the vision encoder better.
The result highlights the learnability of word-referent mapping with short temporally extended episodes.

Interestingly, all three multimodal models perform poorly in classifying the basket class.
This is likely due to the lack of training data for that class.
There exist only around 222 images that belong to the class basket in the Labeled-S dataset.
However, every other class contains images in the range from around 1000 to 10000 samples.
This means that during the pretraining and multimodal contrastive learning, the model is likely to see only very few occurrences of the object basket and even fewer co-occurrences of the object and the utterance.
Since the linear decoding model performs well the basket class, the result could potentially suggest that the contrastive learning model require more data compared to a supervised model.


%% file: sec/6_conclusion.tex
\section{Conclusion}
In this work, using visual experience from the perspective of a child together with child-directed utterances, we explore the learnability of word referent mapping with short-term video and utterance pairs. 
Despite having to evaluate on still images with a video encoder, the multimodal model still achieves competitive performance (-3.639\%) on the CVCL evaluation trial compared to other image-text models. 
Indeed, the result indicates that word referent mapping is learnable from such experiences with contrastive learning using a pretrained masked autoencoder paired with a biologically inspired masking strategy. 
We also evaluate the ability of the video encoder to recognise spatial translation and rotation in the $x$, $y$, and $z$ axes. 
It can generalise the motion of translation and rotation to a diverse range of objects from visual experience alone. 
We demonstrate the importance of learning through temporally continuous experience for the understanding of spatial translation and rotation. 
Learning from still images is not sufficient for spatio-temporal understanding. 
By incorporating visual experience across the time axis, the model is able to learn more than just nouns.